\documentclass[letterpaper, 10 pt, conference]{ieeeconf}  

\IEEEoverridecommandlockouts                              

\usepackage{amssymb}
\usepackage{booktabs}
\usepackage{wrapfig}
\usepackage{amsmath}
\usepackage[table]{xcolor}
\usepackage{graphicx}
\usepackage{subcaption}
\usepackage{booktabs}
\usepackage{multirow}
\usepackage{float}
\usepackage{tabularx}
\usepackage{graphicx}
\usepackage{caption}
\usepackage{cuted}
\usepackage{hyperref}

\title{NavGSim: High-Fidelity Gaussian Splatting Simulator for Large-Scale Navigation}

\begin{document}
\author{%
    \begin{minipage}{\linewidth}
    \centering
    \vspace{3pt}
	Jiahang Liu$^{1,2,*}$ ~
	Yuanxing Duan$^{2,*}$ ~
    Jiazhao Zhang$^{1,2,*}$ \\
	Minghan Li$^{2}$ ~
    Shaoan Wang$^1$ ~
    Zhizheng Zhang$^{2,3,\dagger} ~$
    He Wang$^{1,2,3,\dagger}$ 
    \end{minipage}
	\\ 
    \begin{minipage}{\linewidth}
    \centering
    \vspace{3pt}
    \begin{tabular}{c}
    \normalsize{$^1$Peking University~
	$^2$Galbot~
    $^3$BAAI} \\
    \end{tabular}
    \end{minipage}
    \\
    \begin{minipage}{\linewidth}
    \centering
    \end{minipage}
\thanks{
$^*$ Equal Contribution, $^{\dagger}$ Equal Advising
}
}

\maketitle
\thispagestyle{empty}
\pagestyle{empty}

\begin{strip}
  \centering
  \vspace{-6em} 
  \includegraphics[width=\textwidth, trim=0 110 0 110, clip]{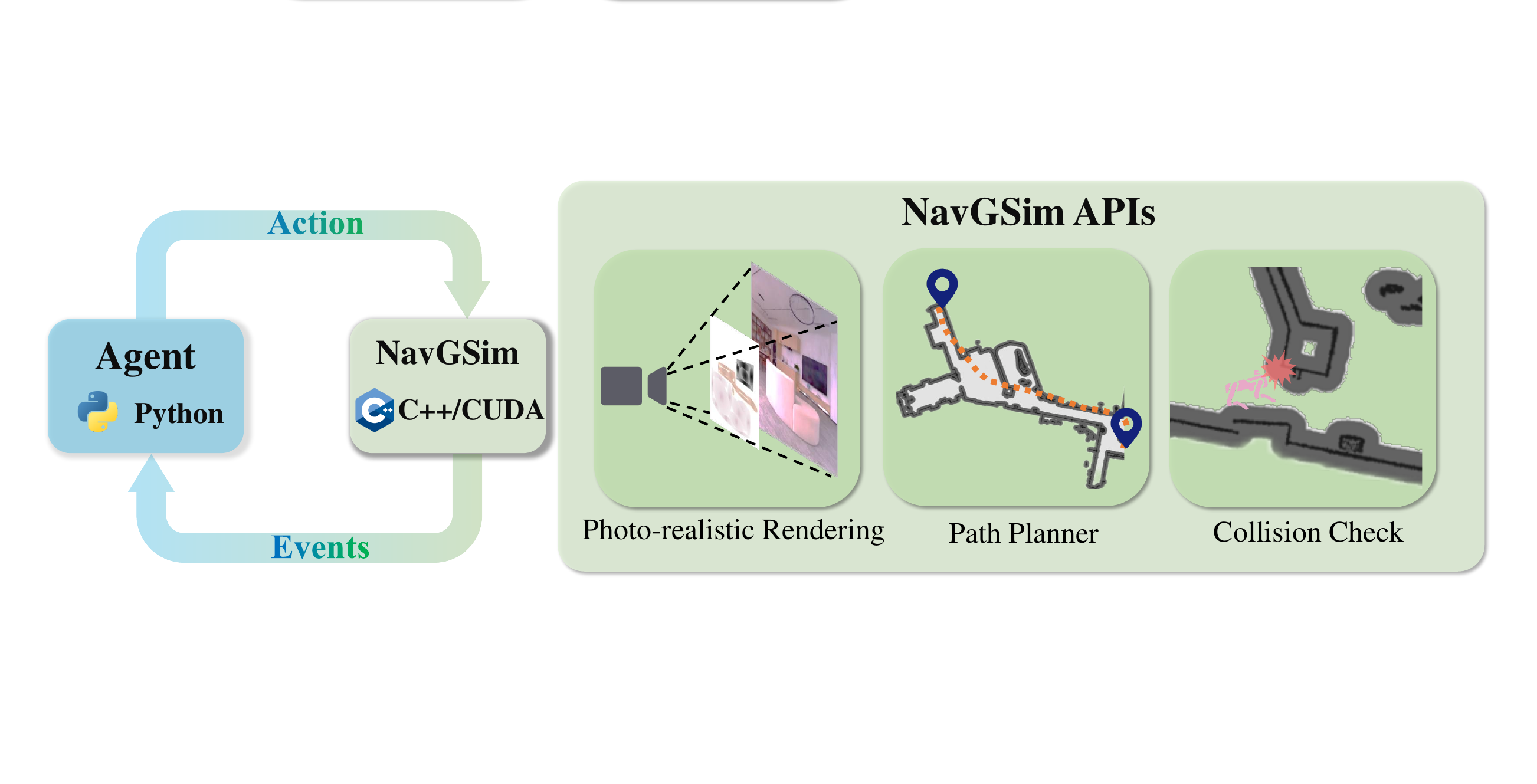}
  \captionof{figure}{\textbf{NavGSim}'s agent-simulator architecture. Users interact with the system through an upper-level python API interface, which provides access to the core functionalities of NavGSim.}
  \label{fig:teaser}
\end{strip}

\begin{abstract}
Simulating realistic environments for robots is widely recognized as a critical challenge in robot learning, particularly in terms of rendering and physical simulation. This challenge becomes even more pronounced in navigation tasks, where trajectories often extend across multiple rooms or even entire floors.
In this work, we present NavGSim, a Gaussian Splatting-based simulator designed to generate high-fidelity, large-scale navigation environments. Built upon a hierarchical 3D Gaussian Splatting framework, NavGSim enables photorealistic rendering in expansive scenes spanning hundreds of square meters. To simulate navigation collisions, we introduce a Gaussian Splatting-based slice technique that directly extracts navigable areas from reconstructed Gaussians. Additionally, for ease of use, we provide comprehensive NavGSim APIs supporting multi-GPU development, including tools for custom scene reconstruction, robot configuration, policy training, and evaluation. To evaluate NavGSim’s effectiveness, we train a Vision-Language-Action (VLA) model using trajectories collected from the NavGSim and assess its performance in both simulated and real-world environments. Our results demonstrate that NavGSim significantly enhances the VLA model’s scene understanding, enabling the policy to handle diverse navigation queries effectively. \textit{NavGSim is publicly available at: https://github.com/2003jiahang/NavGSim}

\end{abstract}

\section{Introduction}
\label{sec:intro}


Simulating realistic environments is widely regarded as a key component of robot learning\cite{argall2009survey,rusu2017sim}, providing a playground for robots to acquire diverse skills. Existing robotics simulators cover a broad range of applications, including locomotion\cite{margolis2024rapid, yang2022fast}, manipulation\cite{kroemer2021review,liu2021deep,shao2021concept2robot}, and navigation\cite{xiao2022motion,zhu2021deep}. Notably, unlike locomotion and manipulation tasks, navigation tasks are distinguished by their requirement for high-fidelity, large-scale scene environments—often spanning multiple rooms or even an entire floor. To simulate such large-scale environments, existing navigation simulators (e.g., Habitat~\cite{savva2019habitat} and AI2-THOR~\cite{kolve2017ai2}) rely on reconstructed or synthetic meshes. However, these approaches often result in unrealistic visual rendering.
While artists can construct highly realistic scenes in game engines like Unreal or Unity, creating large-scale environments still requires significant effort.

In recent years, significant advancements have been made in achieving realistic rendering in navigation simulators through the application of 3D Gaussian splatting (3DGS)~\cite{kerbl20233d}. 3DGS, which provides high-quality visual rendering, has gained  traction in various domains, including navigation, due to its ability to easily reconstruct real-world environments by moving a camera and LiDAR around the environment.Recent works ~\cite{he2025from,deng2025hierarchical,ong2025gaussian,honda2025gsplatvnm,ong2025atlas,ress20253d} have leveraged 3DGS to enhance the fidelity and scalability of navigation tasks. 

Building on these innovations, we proposed NavGSim, a 3D Gaussian Splatting-based simulator designed for high-fidelity, large-scale navigation tasks. Our method extends hierarchical 3D Gaussian Splatting techniques\cite{kerbl2024hierarchical} and incorporates Gaussian-based collision simulation.
Specifically, NavGSim organizes 3D Gaussians into a multi-level hierarchical structure, enabling adaptive level-of-detail (LOD) rendering that dynamically balances visual fidelity and computational efficiency during navigation. To support realistic collision detection, Gaussian ellipsoids are sliced by multiple horizontal planes, and the resulting intersection regions define the object's occupancy representation on each plane.
Consequently, NavGSim converts reconstructed scenes into navigable areas, enabling efficient collision checking and path planning.

Building on NavGSim, we introduce NavGSim-API, a Python library for custom scene reconstruction, robot configuration, policy training, and evaluation. Our API enables researchers to reconstruct scenes from their own posed RGB images. The resulting 3D Gaussian Splatting scenes are fully compatible with NavGSim and support core navigation simulator functionalities, as demonstrated in prior work \cite{savva2019habitat,kolve2017ai2}. 

To evaluate the advantages of NavGSim over existing mesh-based navigation simulators, we fine-tune a vision-language-action (VLA) model~\cite{zhang2024uni} using navigation samples collected directly from NavGSim. We gather 4.5K navigation samples to reach 9 annotated targets. Our results show that the fine-tuned VLA model achieves superior navigation performance in NavGSim and demonstrates strong generalizability in real-world environments by effectively interpreting environmental cues to complete navigation tasks. We hope NavGSim will benefit the research community by enabling the construction of customized navigation simulators for diverse applications.

\section{Related Works}
\label{sec:related}

\textbf{Navigation Simulator.} Navigation simulators\cite{koenig2004design,kolve2017ai2,savva2019habitat,beattie2016deepmind,yan2018chalet,puig2018virtualhome,gao2019vrkitchen,xiang2020sapien,gan2020threedworld} play a crucial role in embodied AI and robot learning by offering controllable environments for navigation tasks such as Vision-and-language navigation\cite{anderson2018vision,krantz2020beyond}, Object Goal Navigation\cite{chaplot2020object,majumdar2022zson}, Embodied Question Answering\cite{
das2018embodied,majumdar2024openeqa} and more. Classic simulators such as Gazebo \cite{koenig2004design}, AI2-THOR \cite{kolve2017ai2} and Habitat \cite{savva2019habitat} provide various levels of realism and interactivity. Note that, the scene environments of navigation simulators often span across multi rooms or one floor. To balance the rendering quality and efficiency, existing navigation simulators adopt conventional rendering techniques\cite{delingette1999general,terzopoulos1991sampling} with reconstrued meshes\cite{ramakrishnan2021habitat,chang2017matterport3d,xia2018gibson} or synthetic meshes\cite{kolve2017ai2,eu2002repair}, which may limit visual fidelity. In contrast, NavGSim leverages 3D Gaussian Splatting, enabling more realistic and scalable simulations.

\textbf{Gaussian Splatting for Robots.} 3D Gaussian Splatting (3DGS) \cite{kerbl20233d} has recently emerged as a compelling alternative to traditional methods like NeRF\cite{mildenhall2021nerf}. It models a scene with a set of 3D Gaussian kernels that jointly encode density, color, and opacity. As a significant advancement in 3D reconstruction, 3DGS has high-quality reconstruction results while keeping real-time rendering performance, offering a powerful tool for bridging real-world data to simulated environments, i.e., Real2Sim\cite{hahn2019real2sim,robinson2009citysim}. For navigation tasks, the required reconstructed environments often span multiple rooms or even entire floors. Although numerous previous works \cite{ren2024scube, liu2024citygaussian, liu2024citygaussianv2, guo2025utilgen} have addressed large-scale scene reconstruction, they cannot be directly applied to navigation tasks, which involve complex layouts with diverse geometries and heavy occlusions. To address this gap, Hierarchical 3D Gaussian Splatting (H3DGS) \cite{kerbl2024hierarchical} proposes a hierarchical region-wise reconstruction framework that spatially decomposes the environment and renders it at multiple levels of detail. This design supports efficient and high-fidelity simulation over large, complex spaces. 

 \begin{figure}[h]
    \centering
    \includegraphics[width=\linewidth, trim=0 20 30 5, clip]{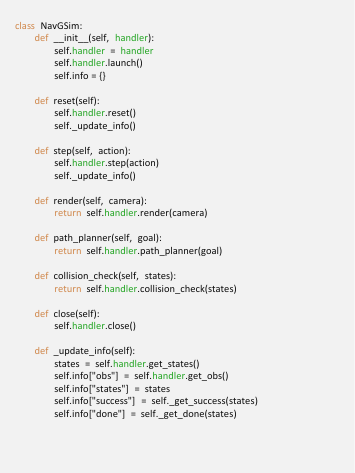}
    \caption{NavGSim's sample code}
    \label{fig:code}
\end{figure}
\vspace{-5pt}
\section{NavGSim Platform}
\label{sec:navsim}

\subsection{NavGSim Architecture}

\begin{figure*}[ht]
    \centering
    \includegraphics[width=1.0\linewidth,trim=0 0 0 0,clip]{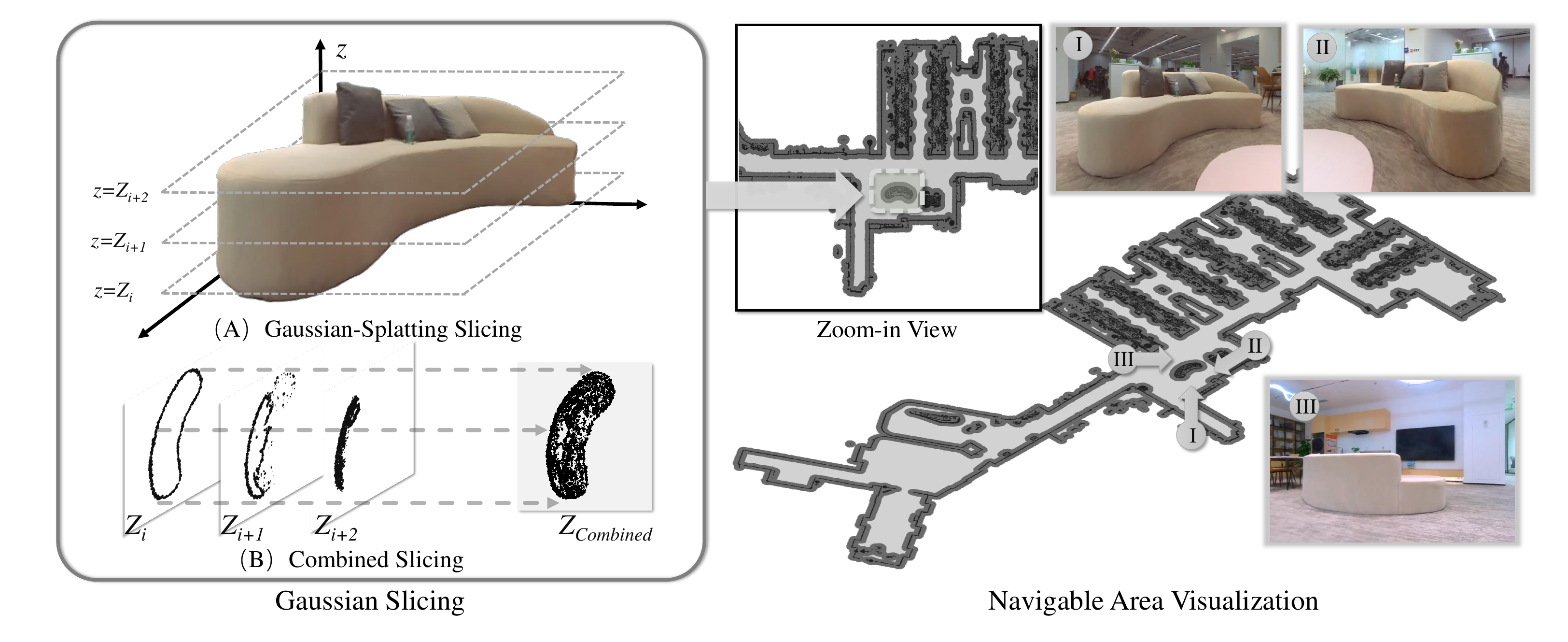}
    \caption{\textbf{Illustration of Gaussian Slicing for Navigable Areas}. The left image shows the 3D Sofa model projected onto multiple horizontal Z-planes ($Z_i$, $Z_{i+1}$, $Z_{i+2}$) at fixed height intervals. Each plane contains a 2D Gaussian slice that represents the projection of the Sofa at that particular height. These slices are combined to form a 2D collision map, with the right image showing the entire scene's 2-D occupancy map that serves as geometric prior information for downstream tasks such as
    navigation cost-map construction or collision checking.}
    \label{fig:slice}
    \vspace{-10pt}
\end{figure*}
Fig.~\ref{fig:teaser} illustrates NavGSim, a modular and scalable simulator for large-scale navigation tasks with photo-realistic rendering path planning, and collision detection. These capabilities are accessible through a simple Python API, supporting the entire pipeline from scene reconstruction to embodied agent training. The usage of APIs is illustrated in Code~\ref{fig:code}. Together, these components form a unified architecture that balances visual realism, navigational reasoning, and ease of use.

NavGSim leverages a Hierarchical 3D Gaussian Splatting~\cite{kerbl2024hierarchical} (H3DGS) engine Sec.~\ref{sec:gaussian_splatting}
 that models indoor scenes as a hierarchy of anisotropic 3D Gaussians. This enables high-quality, real-time rendering from arbitrary viewpoints using only camera intrinsics and extrinsics. To support navigation, we further introduce a planar GS-slice technique Sec.~\ref{sec:collision} that extracts floor-level traversable regions directly from the 3D Gaussian field. This slice provides a dense and accurate approximation of collision volumes, enabling fast and efficient collision check and path planning during simulation.
 Lastly, to facilitate scene annotation, we developed an interactive visualization interface, allowing users to browse the reconstructed scene and label specific objects during their exploration.

\subsection{Hierarchical 3D Gaussian Splatting scenes}\label{sec:gaussian_splatting}

\begin{figure*}[t]
    \centering
    \includegraphics[width=1.0\linewidth, trim=0 10 0 20, clip]{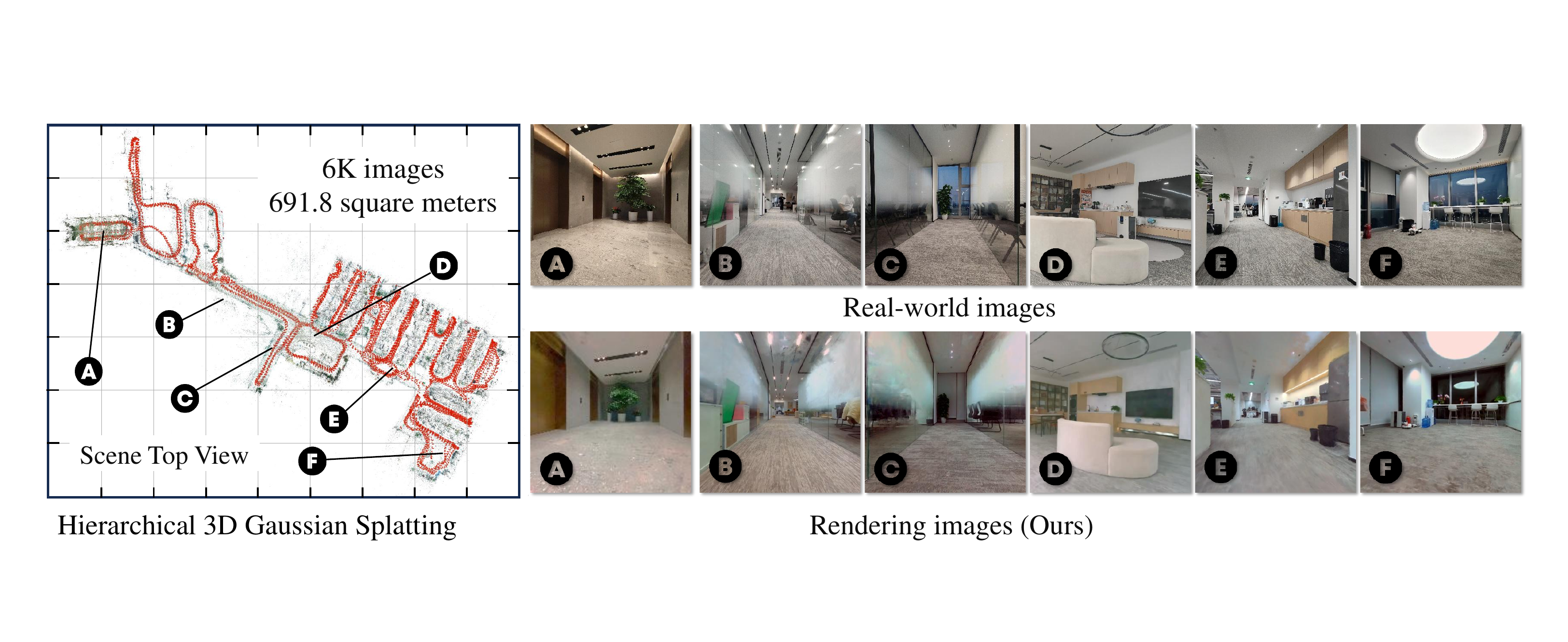}
    \vspace{-2.5em}
    \caption{\textbf{Visualisation of reconstructed Gaussian Splatting and rendering examples.} (Left) We illustrate the process of subdividing the scene into spatial chunks using H3DGS for efficient reconstruction. (Right) We compare rendering results of NavGsim with real-world image at seven distinct viewpoints (A-F).}
    \label{fig:reconstructed}
    \vspace{-5pt}
\end{figure*}

NavGSim utilizes an H3DGS representation for scene modeling and real-time rendering, which is built around a tree-based structure where each node corresponds to a spatial region. Leaf nodes encode fine-grained visual details using dense sets of anisotropic Gaussians, while internal nodes aggregate and abstract these details into coarser levels of representation. This hierarchical structure supports Level-of-Detail (LOD) rendering by dynamically selecting Gaussians from different levels and smoothly interpolating transition regions. To handle large scenes, such as entire floors or multi-room layouts, we adopt a divide-and-conquer strategy. The input image set is divided into spatial chunks, each optimized independently. After local models are trained, a global hierarchy is built by merging and optimizing the chunks, including node cleanup and alignment. We show our reconstruction results in Fig.~\ref{fig:reconstructed}. This approach enables NavSim to process scenes spanning hundreds to thousands of square meters using multiple GPUs in parallel. Ensuring both rendering speed and quality, this 3DGS-based representation also provides the foundation for collision detection in the navigation simulator.

\begin{figure*}[ht]
    \centering
    \includegraphics[width=1.0\linewidth,trim=0 190 0 0, clip]{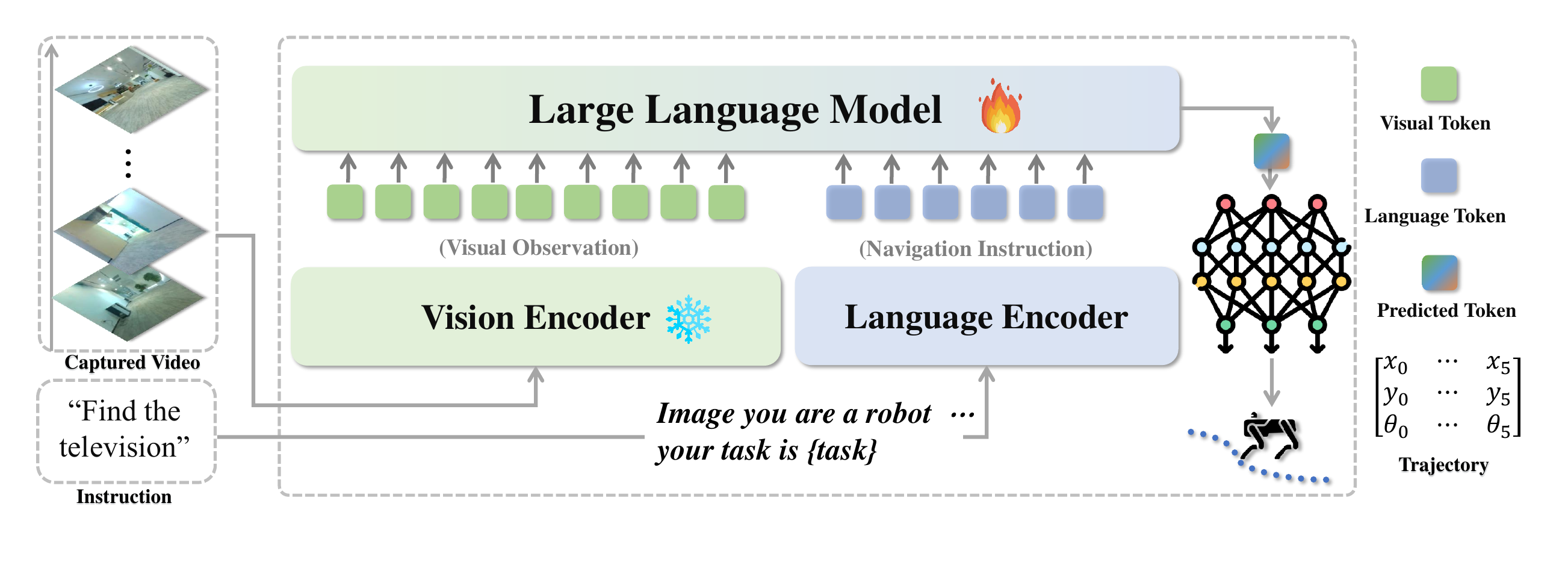}
    \caption{\textbf{VLA-NavGSim architecture}. Our VLA model takes multiple frames of images and language instructions as input. The images are processed by a Vision Encoder to extract visual features. The resulting visual tokens, along with the language tokens, are fed into the LLM to generate a predicted token. This token is then passed through a lightweight MLP to generate a trajectory.}
    \label{fig:VLA}
    \vspace{-10pt}
\end{figure*}

\subsection{GS-based Navigation collision simulation}\label{sec:collision}

To support efficient and realistic collision detection, NavGSim uses a multi-slice Gaussian projection method, which leverages the continuous and probabilistic nature of 3D Gaussian fields, while avoiding the problem of directly projecting 3D ellipses to 2D. Inspired by 4D-Rotor-GS \cite{duan20244d}, which slices 4DGS by different time planes, we define a set of horizontal slicing planes at fixed height intervals (e.g., every 10   cm in the robot's height range) parallel to the floor. Each plane intersects the 3D Gaussian field, and for each Gaussian component, we analytically compute its sliced 2D Gaussian based on its position and 3D covariance. This slicing process is illustrated in Fig.~\ref{fig:slice}.

Specifically, we define the description of 3D Gaussian:

\noindent
\begin{equation}
G_{3D}\!\left ( \mathrm{x} \right ) 
= e^{-\tfrac{1}{2} \left ( \mathrm{x}-\mu \right )^{T} 
   \Sigma_{3D}^{-1} \left ( \mathrm{x}-\mu \right )},
\end{equation}

\begin{equation}
\Sigma_{3D}^{-1} =
\begin{pmatrix}
  c_{xx} & c_{xy} & c_{xz} \\
  c_{xy} & c_{yy} & c_{yz} \\
  c_{xz} & c_{yz} & c_{zz}
\end{pmatrix},
\end{equation}

where \begin{math}\mathrm{x}=\left ( x, y, z \right )\end{math}. Then, given a height z, the sliced 2D Gaussian is obtained as (detailed derivation in the Supplementary Material):
\vspace{-0.5pt}
\begin{equation}
\label{eq:2dgs}
G_{2D}\left ( \mathrm{x}, z \right ) = e^{-\frac{1}{2} \lambda\left ( z-\mu_{z} \right )^{2}} e^{\left [ \mathrm{x}-\mu\left( z\right) \right ]^{T}\Sigma_{2D}^{-1} \left [ \mathrm{x}-\mu\left( z\right) \right ]},
\end{equation}
where \begin{math}\mathrm{x}=\left ( x, y \right )\end{math},

\begin{equation}
\left\{
\begin{matrix}
\alpha c_{xx} + \beta c_{xy} = c_{xz} \\
\alpha c_{xy} + \beta c_{yy} = c_{yz}
\end{matrix},
\right. 
 \qquad
\Sigma_{2D}^{-1} = 
\begin{pmatrix}
  c_{xx} & c_{xy} \\
  c_{xy} & c_{yy} 
\end{pmatrix},
\end{equation}

\begin{equation}
\lambda = c_{zz} - \alpha^{2} c_{xx} - \beta^{2} c_{yy} - 2\alpha\beta c_{xy},
\end{equation}

\vspace{-10pt}
\begin{equation}
\mu \left ( z \right ) = \left ( \mu _x, \mu _y \right ) ^{T} - \left (  z - \mu_z\right ) \left ( \alpha , \beta \right )^{T}.
\end{equation}

In Equation \ref{eq:2dgs}, we introduced a decay term, denoted as \begin{math}\lambda\left ( z-\mu_{z} \right )^{2}\end{math}, which determines the size of the 2D Gaussian slice. When \( z = \mu_z \), the maximum 2D slice of the Gaussian is achieved, and as \( z \) moves further away from \( \mu_z \), the slice gradually shrinks. Simultaneously, \( \mu(z) \) also varies with \( z \), influencing the position of the 2D Gaussian.

Although 3D Gaussian Splatting serves as a probabilistic representation and cannot precisely capture the volume, due to the small size of our Gaussian volumes, we are able to limit the error in the collision volume to within a 5 cm range."

\section{NavGSim for VLA}
\label{sec:vla}

\subsection{VLA architecture}

We extend an existing navigation Vision-Language-Action (VLA) model called Uni-NaVid~\cite{zhang2024navid}, with trajectory output.
Uni-NaVid is a unified video-language model tailored for long-horizon embodied navigation, which fuses visual observations and natural language instructions to predict agent actions. In this work, we adapt and extend Uni-NaVid to enable continuous trajectory prediction, which is essential for realistic navigation in high-fidelity environments.

\textbf{Visual Encoding and Token Compression.}
The model processes a sequence of egocentric RGB frames and a navigation instruction to enable effective multi-modal inference. Each frame is first encoded by an EVA-CLIP\cite{sun2024eva} vision encoder, which divides it into $16 \times 16$ non-overlapping patches, producing 256 visual tokens. To ensure computational efficiency during long-horizon navigation, we adopt the online token merging technique from Uni-NaVid, which temporally organizes visual tokens into three categories: current tokens from the most recent frame with the highest spatial resolution, short-term tokens from the past 64 frames with grid pooling, and long-term tokens from earlier frames with aggressively pooling and token merging based on a cosine similarity threshold. These compressed visual tokens are then projected into the language embedding space via a Cross-Modality Projector (2-layer MLP) to match the latent space of the Vicuna-7B LLM\cite{chiang2023vicuna}. The navigation instruction is tokenized and embedded using the same LLM. Finally, the visual and language tokens are concatenated into a unified input sequence: \texttt{\{Long\_term\_tokens\} \{Short\_term\_tokens\} \{Current\_tokens\} <NAV> \{Instruction\}}
, allowing the LLM to jointly attend to both visual observations and instructions.

\textbf{Modified Output Representation.}
Unlike the original Uni-NaVid design, which outputs discrete action tokens, inspired by~\cite{shah2023gnm}, we reformulate the prediction target to suit continuous trajectory planning. Specifically, the LLM performs only a single autoregressive step, and we extract the hidden state from its final layer as the condition, which is subsequently passed through an MLP-based action head to predict the agent’s future trajectory in the form: $(x, y, \theta) \in \mathbb{R}^3$, where $(x, y)$ and $\theta$ denote the agent’s position and yaw angle in its current coordinate frame, respectively. Compared to the discrete action space used in Uni-NaVid, our continuous trajectory output enables smoother and more realistic motion, better aligning with the requirements of large-scale real-world or simulated environments, where discretization may lead to unnatural movements or overshooting.

\subsection{Data Collection}\label{sec:data}

We collect a diverse set of 4,500 navigation trajectories within the NavGSim simulator. Each trajectory starts from a randomly sampled navigable position and heads toward a semantically meaningful goal, ensuring broad coverage of spatial configurations and behaviors.



\textbf{Targets Annotation.} Within the reconstructed environment, we manually annotate 9 semantic landmarks (\textit{e. g.}, elevator, television, meeting room, etc.) as navigation goals. For each goal, we collect 200 expert trajectories generated by a shortest-path planner. Each trajectory starts from a randomly sampled position and rotation and ends at the selected specified landmark, ensuring a wide variety of spatial configurations and path topologies. To address the compounding error problem in policy inference, we employ the DAgger algorithm~\cite{ross2011reduction} to iteratively augment the dataset. In each round of DAgger, the agent navigates using its own policy, while an expert policy provides supervision at each step. In this work, we conduct two rounds of DAgger, collecting 150 additional trajectories per landmark in each round, resulting in a total of 2,700 trajectories. This significantly enhances the robustness and diversity of the training dataset.

\textbf{Imitation Data Collection}. Initially, the dataset is constructed using trajectories generated by a shortest-path planner operating under idealized simulation conditions. Although this supervised data helps the model learn fundamental navigation priors, we observe a significant drop in performance during inference. Specifically, the model tends to accumulate small prediction errors over time, which eventually lead to navigation failure—particularly in long-horizon tasks or cluttered environments.

\textbf{Dagger Data Collection}. To address this issue, we incorporate Dataset Aggregation (DAgger)\cite{ross2011reduction}, a technique that iteratively refines the model by collecting data under its own policy. During rollout, the agent executes its predicted actions, and trajectory collection is terminated under one of the following conditions:
(1) the agent collides with the environment for n consecutive steps (we set n=8);
(2) the number of inference steps exceeds a predefined budget;
(3) the agent successfully reaches the goal.
For each collected trajectory, the expert policy is queried to provide corrective supervision when the agent's behavior diverges. This enables the model to learn from its own failure cases, improving robustness in out-of-distribution states and enhancing recovery capabilities.


Together, these strategies yield a training corpus that combines expert planning, DAgger-augmented behavior, and multi-modal supervision. This hybrid approach equips the VLA model with both accurate motion prediction and high-level scene comprehension, enabling reliable long-horizon navigation in complex environments.

\section{Experiments}
\label{sec:exp}

\begin{figure*}[ht]
    \centering
    \includegraphics[width=1.0\linewidth, clip]{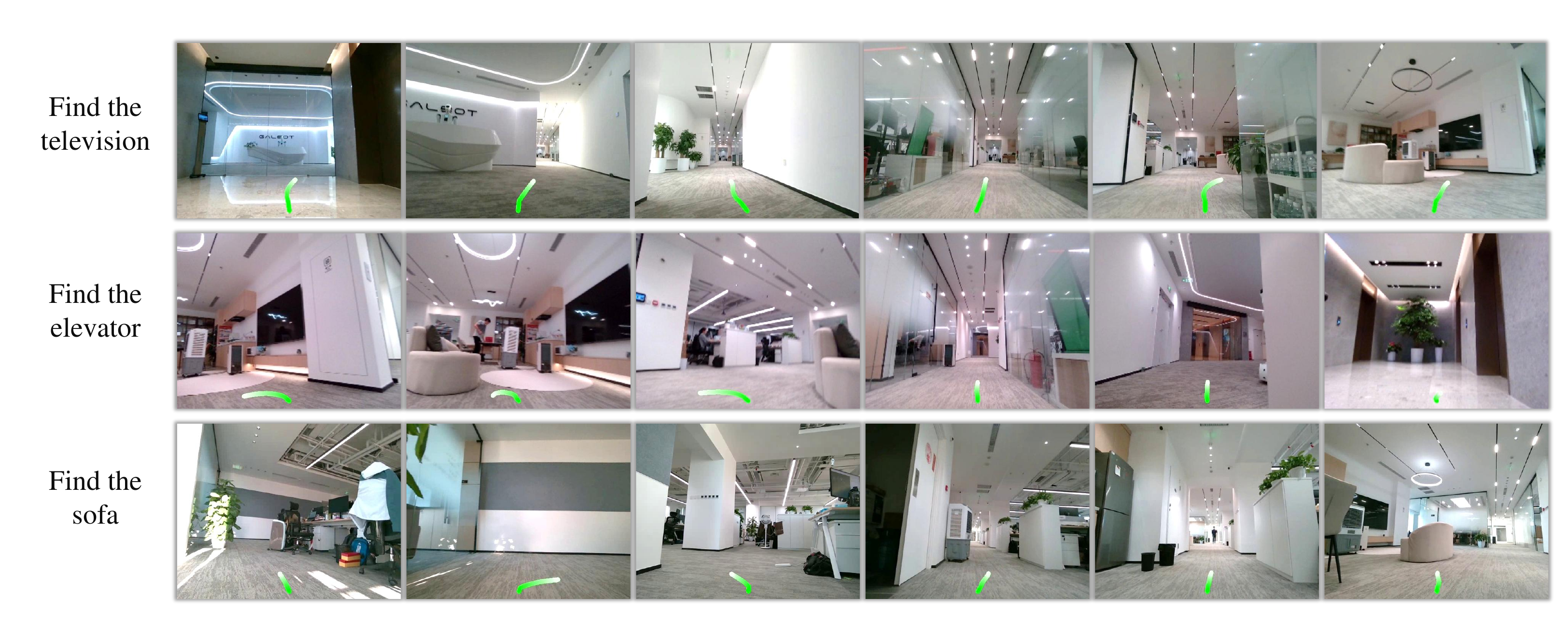}
    \caption{\textbf{Visualization of real-world results.} Visualization First-person perspective display of real-world testing.}
    \label{fig:real}
    \vspace{-1em}
\end{figure*}
We evaluate the effectiveness of NavGSim in both simulation and real-world environments through comprehensive experiments. Our goal is to validate whether the trained VLA model can generalize to novel start positions, maintain high success rates during long-horizon navigation, and transfer its policy effectively from simulation to the real world.

\subsection{Experimental Setup}


\textbf{Gaussian Splatting Scene Construction.} To evaluate NavGSim, we constructed a realistic large-scale indoor simulation environment of an office factory, covering approximately 700~m$^2$. We used a handheld LiDAR-RGB camera device to perform a walk-through scan of the entire scene, capturing multi-view images and LiDAR point clouds along natural human trajectories. These data were then processed using our H3DGS pipeline to reconstruct a photorealistic 3D environment.

\textbf{Training Details of VLA.} We train the VLA model using a combination of supervised learning and DAgger-augmented imitation data, as described in Sec.~\ref{sec:data}. To further improve generalization, we also incorporate video question-answering (VQA) data from publicly available datasets\cite{azuma2022scanqa,chen2024panda,li2024llama} into the training process. 
During training, the vision encoder (EVA-CLIP~\cite{sun2024eva}) is frozen to preserve its pre-trained visual features, while the large language model (Vicuna-7B~\cite{chiang2023vicuna}) is fully fine-tuned along with the cross-modal projection layers and the trajectory regression head. Training is conducted on a cluster of 16 NVIDIA H800 GPUs for 20 hours, totaling 320 GPU-hours.

\textbf{Real-World Deployment and Evaluation.} We deploy our NavGSim-trained VLA agent on a Unitree Go2 quadruped robot in office environments, using a front-facing RGB camera to stream egocentric vision via WebSocket to an NVIDIA A100 server. The purely simulated-trained model fuses real-time visual data with language instructions to generate navigation waypoints, achieving 8 Hz closed-loop control through the robot's onboard planner (Unitree extension dock). This modular architecture decouples semantic planning from low-level execution while maintaining responsiveness, with physical deployment demonstrations captured in Fig.~\ref{fig:real}.

\textbf{Metrics.} During evaluation, we randomly sample initial positions and orientations and measure the success rate of navigation to the 9 annotated landmarks and two unannotated novel objects. A trajectory is considered successful if the agent stops within 1.5 meters of the target. We use two metrics: Success Rate (SR) and Distance to Goal (DTG).

\subsection{Performance Analysis}

\begin{figure}[htbp]
    \centering
    \includegraphics[width=1.0\linewidth, clip]{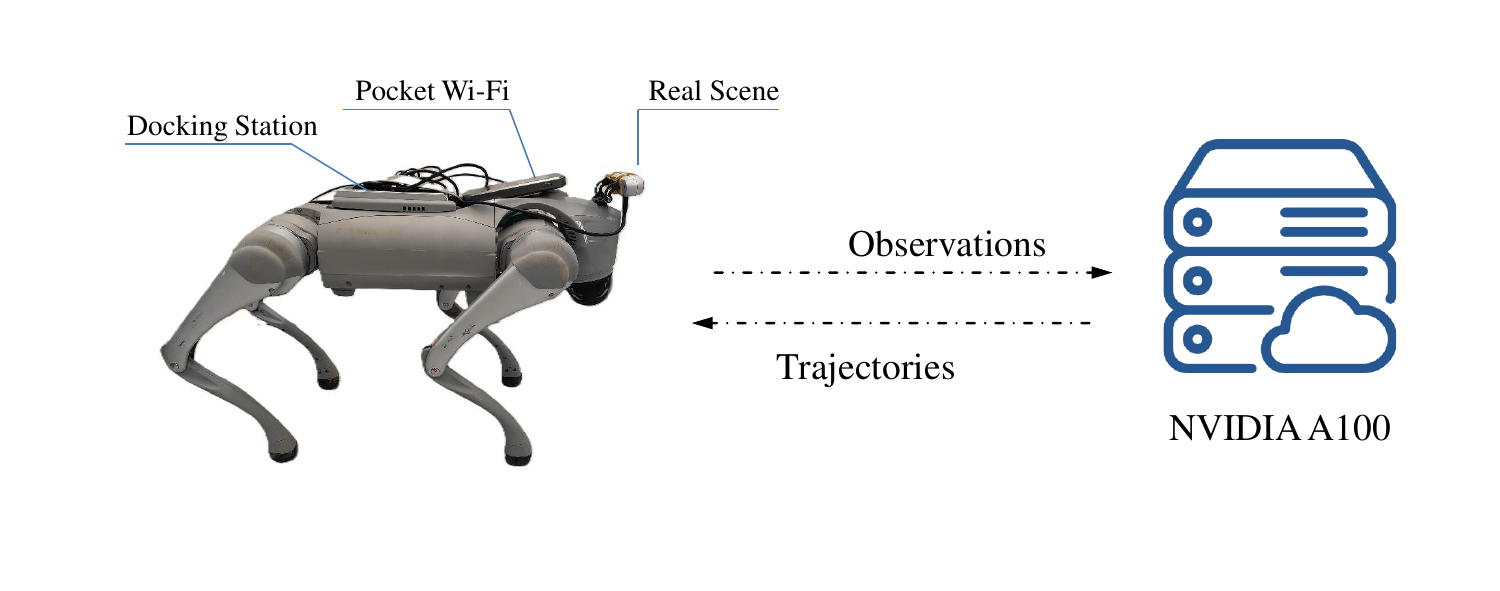}
    \caption{\textbf{Real machine deployment architecture} The agent interacts with the server via the Internet.}
    \label{fig:platform}
    \vspace{-10pt}
\end{figure}

\begin{table*}[t]
  \centering
  \renewcommand{\arraystretch}{1.1}
  \setlength{\tabcolsep}{16pt} 
    \begin{tabular}{clccc}
      \toprule
      \multirow{2}{*}{\textbf{Target Category}} & \multirow{2}{*}{\textbf{Target}} & \multicolumn{3}{c}{\textbf{H3DGS / 3DGS }} \\
      \cmidrule(lr){3-5}
      & & Sim SR(\%)~$\uparrow$ & Sim DTG(m)~$\downarrow$ & Real SR(\%)~$\uparrow$ \\
      \midrule
      \multirow{10}{*}{Annotated Targets}
      & \texttt{Pantry}             & 96.7 / 83.3  & 0.61 / 1.75  & 66.7 / 0.00 \\
      & \texttt{Elevator}           & 98.3 / 83.3  & 0.58 / 2.73  & 71.4 / 16.7 \\
      & \texttt{Restroom}           & 96.7 / 86.1  & 1.77 / 2.10  & 73.3 / 23.1 \\
      & \texttt{Television}         & 100 / 91.6 & 0.39 / 1.28  & 73.3 / 18.1 \\
      & \texttt{Laboratory}         & 100 / 83.3 & 0.30 / 1.64  & 75.0 / 18.1 \\
      & \texttt{Meeting Room I}     & 100 / 80.6 & 0.88 / 2.62  & 80.0 / 23.1 \\
      & \texttt{Meeting Room II}    & 100 / 77.8 & 0.50 / 3.82  & 63.0 / 15.4 \\
      & \texttt{Reception Room}     & 98.3 / 72.2  & 1.21 / 4.29  & 63.3 / 9.10 \\
      & \texttt{Water Dispenser}    & 95.0 / 69.4  & 1.92 / 5.42  & 57.1 / 0.00 \\
      & \texttt{Avg}                & \colorbox{red!25}{98.3} / 80.8  & \colorbox{red!25}{0.91} / 2.85  & \colorbox{red!25}{69.2} / 13.7 \\
      \midrule
      \multirow{3}{*}{Novel Targets}
      & \texttt{Table}              & 78.3 / 52.8  & 3.91 / 7.62  & 68.8 / 16.7 \\
      & \texttt{Sofa}               & 83.3 / 55.6  & 2.56 / 6.97  & 70.4 / 0.00 \\
      & \texttt{Avg}                & \colorbox{red!25}{80.8} / 54.2  & \colorbox{red!25}{3.23} / 7.29  & \colorbox{red!25}{69.6} / 8.35 \\

      \bottomrule
    \end{tabular}
  \caption{\textbf{Comparison of different rendering techniques}. We report navigation performance across seen and unseen targets in the simulator and real-world environments. }
  \label{tab:nav_results}
\end{table*}

\textbf{Performance on simulator and real-world environments.}. We evaluate the navigation performance of the NavGSim-trained VLA model across both simulated and real-world environments.
Table~\ref{tab:nav_results} summarizes the navigation success rates and stop distances across seen and unseen targets in the simulation environment.
The model demonstrates strong performance on seen landmarks such as \textit{Television} (100.0\%) and \textit{Laboratory} (100.0\%), achieving precise and efficient trajectories as indicated by low stop distances. Furthermore, it generalizes effectively to unseen targets, with success rates of 78.3\% on \textit{Table} and 83.3\% on \textit{Sofa}, suggesting robust spatial understanding and goal generalization beyond the training distribution.
In real-world deployment, despite the presence of sensor noise and domain discrepancies, the trained model maintains high success rates on key targets, demonstrating the model's ability to transfer navigation skills learned in simulation to physical environments. See the Supplemental Material for the results video.

\begin{table}
    \centering
    \setlength{\tabcolsep}{10pt} 
    \begin{tabular}{lccc}
    \toprule
    Method & PSNR$\uparrow$ & LPIPS$\downarrow$ & SSIM$\uparrow$  \\
    \midrule
    H3DGS  & 25.396  & 0.274  &  0.866 \\
    3DGS~\cite{kerbl20233d}  & 23.221  & 0.426  &  0.789    \\
    \bottomrule
    \end{tabular}
    \caption{Comparison on different methods}
    \label{tab:ablation_action}
\end{table}

\textbf{Comparison of rendering quality}. Moreover, we compare the training performance in a simulator built on original 3DGS. As shown in Table~\ref{tab:ablation_action}, the rendering quality of the original 3DGS simulator is lower than that built by H3DGS. Although the model performs reasonably well in simulation, its low-quality rendering results in a significant performance gap when transferred to the real world, resulting a sharp decrease (81\%~$\downarrow$) in real-world performance. 
These results confirm the practical feasibility of sim-to-real transfer enabled by the NavGSim framework.

\begin{figure}[tbp]
    \centering
    \includegraphics[width=0.8\linewidth]{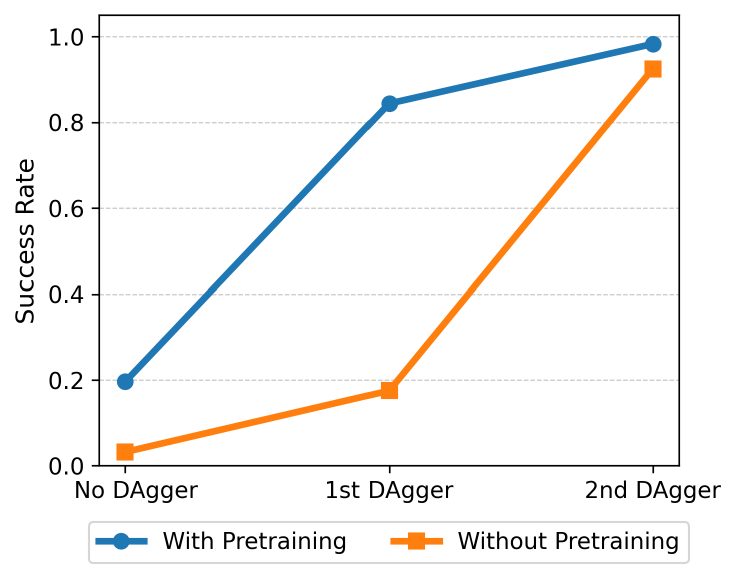}
    \caption{\textbf{Performance on NavGSim with different training strategy}. Success rate comparison with and without pretraining across different training stages.}
    \vspace{-3pt}
    \label{fig:success_rate}
\end{figure}


\textbf{Influence of training strategy}. We compared the progression of the success rate in different training stages for models trained with and without pre-training, as shown in Figure~\ref{fig:success_rate}.
The model utilizing pretraining shows significantly accelerated learning and higher final success rates, achieving over 90\% success after two rounds of DAgger.
In contrast, models trained from scratch exhibit slower improvement and plateau at substantially lower performance levels.
This comparison highlights the critical role of pretraining in improving data efficiency, reducing compounding errors, and enabling more effective long-term navigation policy learning.The pre-trained models leverage previously learned features, allowing them to make faster adjustments and achieve a higher success rate much earlier in the training process.

\section{Conclusions}
\label{sec:conclusion}

In this work, we present NavGSim, a simulator based on Hierarchical Gaussian Splatting for high-fidelity, large-scale navigation. Compared to traditional mesh-based simulators, NavGSim achieves photorealistic rendering and precise collision modeling via the proposed GS-Slice technique. We integrate NavGSim into a real2sim2real navigation pipeline, constructing large-scale environments and adopting an adapted Uni-NaVid architecture with continuous pose regression. Experiments in both simulation and real-world settings demonstrate improved scene understanding for VLA models and effective sim-to-real transfer. By bridging high-quality scene reconstruction and embodied learning, NavGSim enables scalable and generalizable robot navigation.

\section{Limitations}
\label{sec:limitations}

Despite its strengths, NavGSim still has several limitations. First, the hierarchical Gaussian plating pre-processing pipeline is time-consuming, especially for large-scale scenes, which limits the system’s flexibility in rapidly iterating over new environments. Second, the current simulator design does not support dynamic editing, such as interactively adding or removing objects after reconstruction. This constraint makes it less suitable for tasks requiring frequent scene modifications or procedural generation. Addressing these limitations is crucial for enabling broader applicability in real-world robotics and simulation-intensive research workflows. Future work will explore faster scene reconstruction pipelines, dynamic environment modeling, and more lightweight sim2real adaptation strategies to further improve scalability and practicality.


\section{Appendix}
\subsection{Stop indication.}

\begin{figure}[htbp]
    \centering
    \includegraphics[width=0.8\linewidth, clip]{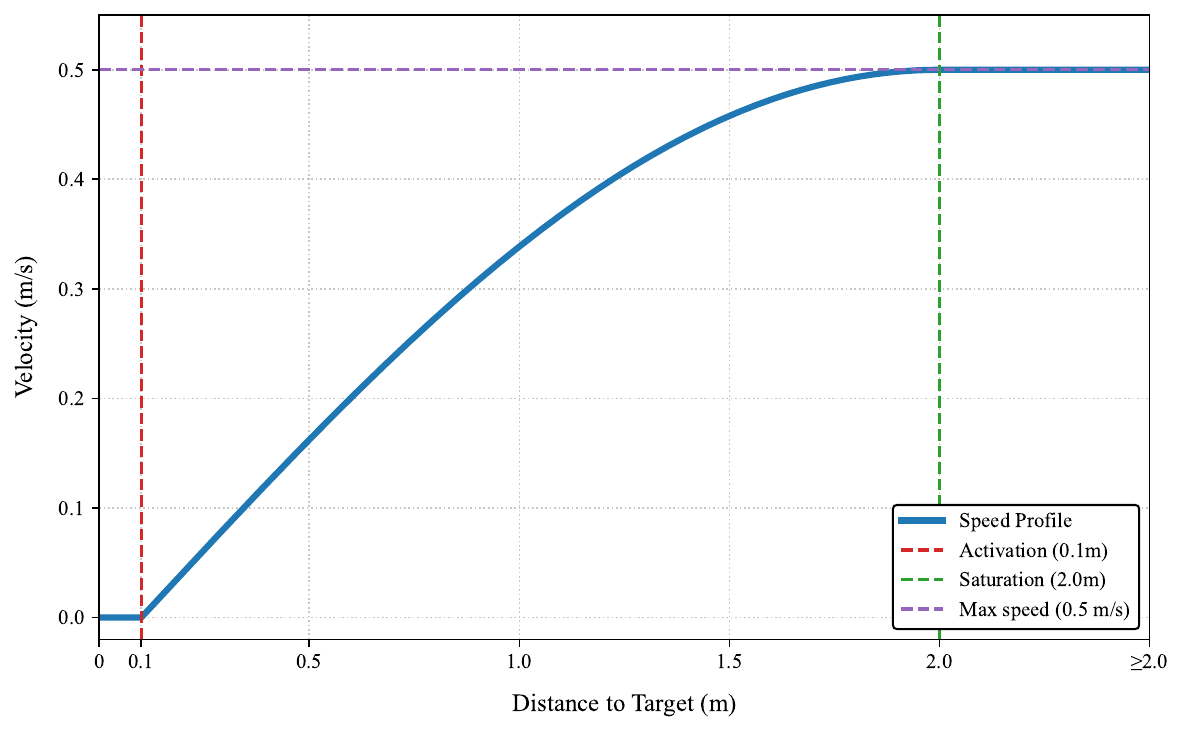}
    \caption{\textbf{Robot Velocity vs Target Distance} }
    \label{fig:speed}
    \vspace{-10pt}
\end{figure}

In our approach, we output continuous trajectories, which, compared to discrete trajectories, provide smoother paths. However, a significant issue arises: we cannot explicitly instruct the model to output a "stop" command. Therefore, an important challenge is how to gradually decelerate the model as it approaches the destination. To address this, we designed a velocity modulation model. When the agent enters a radius of 2 meters from the target, its speed gradually decreases until it reaches a stop within 10 centimeters of the target, as shown in Figure~\ref{fig:speed}. During testing, if the agent's velocity drops to 0.1 m/s, we consider this as the "stop" point and halt the agent at that position.

\subsection{Trajectory length.}

When generating training trajectories, we randomly sample starting points uniformly throughout the environment. Due to the diverse spatial distribution of target locations, the resulting trajectories naturally vary in length. Detailed statistics are presented in Table~\ref{tab:trajectory_lengths}.

\subsection{Detailed derivation of Gaussian Slicing.}
In this section, we provide the details about slicing the 2D Gaussian to obtain the slice at a given height \(z\). The 2D Gaussian is sliced by intersecting it with a plane at height \(z\), and we calculate the 2D center position and 2D covariance after being intercepted by the \(z\)-plane.

First, we have the 3D covariance matrix represented by \(\Sigma_{3D}\) and the rotation \(R_{3D}\).

\[
\Sigma_{3D} = R_{3D} S_{3D}S_{3D}^T R_{3D}^T
\]

\[
\Sigma_{3D}^{-1} = R_{3D} S_{3D}^{-1}(S_{3D}^{-1})^T R_{3D}^T
\]

Then we obtain the inverse of the 3D covariance matrix \(\Sigma_{3D}^{-1}\), expressed as:

\[
\Sigma_{3D}^{-1} = \begin{pmatrix}
  c_{xx} & c_{xy} & c_{xz} \\
  c_{xy} & c_{yy} & c_{yz} \\
  c_{xz} & c_{yz} & c_{zz}
\end{pmatrix}
\]

Since a 3D Gaussian can be expressed as:

\[
G_{3D}(\mathbf{x}) = e^{-\frac{1}{2} \mathbf{x}'^T \Sigma_{3D}^{-1} \mathbf{x}} = e^{-\frac{1}{2} B}
\]

where \(\mathbf{x}' = \mathbf{x} - (\mu_x, \mu_y, \mu_z)\). We get the following:

\[
B = x'^2 c_{xx} + y'^2 c_{yy} + z'^2 c_{zz} + 2 x' y' c_{xy} + 2 x' z' c_{xz} + 2 y' z' c_{yz}
\]

Then, solving for the 2D Gaussian slice at height \(z\), we set:

\[
\begin{split}
B =\;& c_{xx}(x' + \alpha z')^2\\
& + c_{yy}(y' + \beta z')^2 + 2c_{xy}(x' + \alpha z')(y' + \beta y')\\
& + \big(c_{zz} - c_{xx}\alpha^2 - c_{yy}\beta^2 - 2c_{xy}\alpha\beta\big) z'^2
\end{split}
\]

After that we get the equation:

\[
\left\{
\begin{matrix}
\alpha c_{xx} + \beta c_{xy} = c_{xz} \\
\alpha c_{xy} + \beta c_{yy} = c_{yz}
\end{matrix}
\right.
\]

After solving \(\alpha, \beta\), we obtain the position of 2D gaussian sliced at \(z\):

\[
\left\{
\begin{matrix}
\mu_{x}(z) = \mu_{x} - \alpha( z - \mu_{z}) \\
\mu_{y}(z) = \mu_{y} - \beta( z - \mu_{z})
\end{matrix}
\right.
\]

Finnally, the resulting 2D Gaussian slice equation can be written as:

\[
G_{2D}(\mathbf{x}, z) = e^{-\frac{1}{2} \lambda (z - \mu_z)^2} e^{\left[ \mathbf{x} - \mu(z) \right]^T \Sigma_{2D}^{-1} \left[ \mathbf{x} - \mu(z) \right]}
\]

where:

\[
\Sigma_{2D}^{-1} = \begin{pmatrix}
  c_{xx} & c_{xy} \\
  c_{xy} & c_{yy}
\end{pmatrix}
\]

\[
\mu(z) = \left( \mu_x, \mu_y \right)^{T} - (z - \mu_z) \begin{pmatrix} \alpha \\ \beta \end{pmatrix}
\]

\[
\lambda = c_{zz} - \alpha^2 c_{xx} - \beta^2 c_{yy} - 2\alpha\beta c_{xy}
\]

\begin{table}[tbp]
  \centering
  \renewcommand{\arraystretch}{1.3}
  \setlength{\tabcolsep}{12pt} 
  \caption{\textbf{Average trajectory lengths for different target categories.} We report the average navigation trajectory length (in meters) across different indoor target locations.}
  \label{tab:trajectory_lengths}
  \begin{tabular}{l c}
    \toprule
    \textbf{Category} & \textbf{Avg. Trajectory Length (m)} \\
    \midrule
    \texttt{Pantry}             & 19.89 \\
    \texttt{Elevator}           & 26.06 \\
    \texttt{Restroom}           & 24.29 \\
    \texttt{Television}         & 18.29 \\
    \texttt{Laboratory}         & 16.80 \\
    \texttt{Meeting Room I}     & 19.43 \\
    \texttt{Meeting Room II}    & 18.41 \\
    \texttt{Reception Room}     & 26.14 \\
    \texttt{Water Dispenser}    & 30.98 \\
    \midrule
    \textbf{Overall Avg.} & \textbf{22.26} \\
    \bottomrule
  \end{tabular}
\end{table}

\subsection{Real Word Deployment.}


In figure~\ref{fig:platform}, we present the complete robot hardware platform, which consists of the Unitree Go2 quadruped robot, equipped with an Intel RealSense D455 camera, portable Wi-Fi and a docking station. Due to computational constraints, our model is deployed on a remote NVIDIA A100 server. During the test, images captured from the first-person perspective by RealScene are transmitted through portable Wi-Fi through the internet to establish communication with the server. These images are fed as input to the model, which returns the trajectory coordinates (x, y, \(\theta)\). Finally, the trajectory data are converted into robot control signals through a local planning algorithm deployed at the docking station. In addition, to ensure safety during the experiment, we utilize radar data to avoid obstacles.

\newpage

\bibliographystyle{IEEEtran}
\bibliography{IEEEabrv,references}

\end{document}